# Performance Study of Distributed Apriori-like Frequent Itemsets Mining

**Lamine M. Aouad, Nhien-An Le-Khac, and M-Tahar Kechadi**

**School of Computer Science, College of Science,
University College Dublin (UCD)**

**Abstract**  In this article, we focus on distributed Apriori-based frequent itemsets mining. We present a new distributed approach which takes into account inherent characteristics of this algorithm. We study the distribution aspect of this algorithm and give a comparison of the proposed approach with a classical Apriori-like distributed algorithm, using both analytical and experimental studies. We find that under a wide range of conditions and datasets, the performance of a distributed Apriori-like algorithm is not related to global strategies of pruning since the performance of the local Apriori generation is usually characterized by relatively high success rates of candidate sets frequency at low levels which switch to very low rates at some stage, and often drops to zero. This means that the intermediate communication steps and remote support counts computation and collection in classical distributed schemes are computationally inefficient locally, and then constrains the global performance. Our performance evaluation is done on a large cluster of workstations using the Condor system and its workflow manager DAGMan. The results show that the presented approach greatly enhances the performance and achieves good scalability compared to a typical distributed Apriori founded algorithm.

## 1 Introduction

Mining frequent itemsets is at the core of various applications in the data-mining field. The best known such task is the association rules finding. Since its inception, many frequent itemset mining algorithms have been proposed in the literature [1–5], etc. Many of them are related to the Apriori approach. Basically, frequent itemsets generation algorithms analyse the dataset to determine which combination of items occurs together frequently. For instance, considering the commonly known market basket analysis; each customer buys a set of items representing his/her basket. The input of the algorithm is a list of $t$ transactions giving the sets of items, among $I$ items, in each basket. For a fixed threshold support $s$, the algorithm determines which sets of items, of a given size $k$, are contained in at least $s$ of the $t$ transactions or baskets.

Specifically, we are focusing on mining frequent itemsets on distributed datasets over the grid. Indeed, huge amount of datasets are naturally distributed over different geographic locations. Furthermore, grid technologies have recently emerged as a de facto standard for distributed computing. A large amount of scientific communities are using available distributed facilities to share, manage and process large scale datasets and applications. Grid-based approaches are then motivated by the inherent distributed nature and by the challenge of developing scalable solutions for data-mining applications, which are highly computationally expensive and data intensive, taking into account the constraints related to the analysis of these massive distributed datasets on these environments. Many high-level distributed and grid-based knowledge discovery and data-mining systems have been, or are being, proposed leading to the need of effective high-level algorithmic approaches. This article introduces the problem of Apriori-based frequent itemsets mining of large distributed collections of datasets over the grid and presents a well-adapted distributed approach for this purpose, based on both analytical and experimental approaches.

The analytical part of this article has a performance analysis of the Apriori task under a parameterised model, and a network performance model based on LogP. Each local dataset portion has a certain number of candidates at each generation level. This is the main factor that determines how much a classical-distributed implementation is communicating, in addition to the main factor on how much the algorithm will do locally. The worst-case work and communication needed by a classical distribution of the Apriori algorithm can be exponential in the size of the input. Considering the case where every transaction, in every node, contains every item, the algorithm must output and may communicate each subset of the $I$ items, at each level. This can be due to remote support counts collections for global pruning purposes. In this case, the communication steps are completely unnecessary. We will show that local pruning strategies are mostly sufficient and that intermediate global phases do not bring enough useful information, and performance constraining. We will give analytical formulations of the presented approach and a classical Apriori founded distributed algorithm, namely the fast distributed mining (FDM) of association rules algorithm, proposed in [6] and [7].

The experimental part of the article gives performance evaluations and various factors estimations for the performance models. Two large datasets are tested, a synthetic dataset generated using the IBM Quest generation code and a census dataset available from the UC Irvine KDD Archive, and derived from the public use microdata samples (PUMS) dataset. Conclusions from experimental datasets will give understanding of the amount of work done by the Apriori algorithm in both cases, and define bounds of the factors and parameters described in this article. The main factor, leading our reasoning, is how close are the candidate sets at each level corresponding to both approaches. This is also related to the success rate, which is usually high for small values of the level $l$, and low after a given value of $l$, and often drops to zero.

Considering the performance behaviour of the Apriori generation task will then lead to consider the intermediate communication steps as computationally inefficient in comparison to the related overheads which are cumulated at each level. We will show that this process can be done on a single phase at the end of local countings much more efficiently. Based on these observations, the proposed approach has two phases: the local mining phase, and the global counts collection phase. In the first phase, we consider only a local pruning strategy. In the second phase, each node asks for remote support counts of a collection of locally frequent itemsets from the other nodes. This phase is carried out in few passes. All globally frequent itemsets from size 1 to $k$ can be deduced by considering global support counts collections using a top-down search. The overhead related to multiple synchronisation and communication phases in classical approaches can then be highly reduced using a single collection phase with much fewer passes. While our performance study focuses on the Apriori-based distribution, we believe that the key reasoning of this study will hold for many other frequent itemsets generation tasks, since it is partly related to the dataset properties.

The rest of the article is organized as follows. The next section surveys related work in parallel and distributed frequent itemsets mining. In Sect. 3, we give the problem definition, then in Sect. 4, we present the algorithm and the analytical analysis for both the proposed approach and the FDM algorithm. In Sect. 5, we present the experimentation setup. The following section presents the performance evaluation as well as factors bounds related to the proposed models. Finally, concluding remarks are made in Sect. 7.

## 2 Related work

Frequent itemsets generation plays an important role in a range of tasks in data mining including association rules, correlations, causality, episodes, among others. Most of the existing approaches were dedicated to parallel systems using standard message passing communication interfaces. These include, among others, the count distribution (CD) approach in [8], optimized distributed association rule mining (ODAM) [9], the FDM approach [6,7], the CCPD approach in [10] and [11], parallel FP-Growth [12], etc. In this section, we will not be considering approaches that need a global view, the dataset replication, or dataset portion movements since they are not suited for loosely coupled environments.

Most of the following approaches assume that the datasets are horizontally partitioned among processes. The CD algorithm is a straightforward parallelisation of the Apriori algorithm [1]. In order to obtain global counts, local counts are exchanged after each process has generated its local candidate itemsets. This approach was enhanced by the FDM algorithm which generates a smaller number of candidate itemsets. It is intended to reduce communication costs by combining different strategies of local and global prunings for the candidate sets generation. It implements some message optimization techniques that require an assignment function, which can be a hash function, for selecting polling nodes. However, this can generate an extra computational cost and increase communication overhead caused by the global support counts requests from polling nodes when the number of nodes is large. Another Apriori-based variant is proposed by the ODAM algorithm. It intends to minimise the candidate itemset generation cost by eliminating all infrequent items after the first pass to efficiently generate candidate support counts in latter passes, and adopts the FDM pruning technique for intermediate candidate itemsets generation.

In the group of Apriori-based algorithms, we can also cite the CCPD approach [11], targeting shared-memory machines and proposing locality enhancing and false sharing reducing policies, and the distributed decision miner (DDM) approach and its variants [13]. This latter approach introduces a distributed decision protocol in each round in order to determine the set of globally frequent itemsets. It aims to reduce the communication needs and enhance scalability compared to the previously quoted Apriori founded algorithms.

Other distributed approaches are based on different sequential algorithms, such as the FP-Growth approach [12], or the Sampling algorithm [14]. A combination of the latter algorithm and the DDM approach, called D-Sampling, was proposed in [15]. In addition, many approaches are dedicated to maximal frequent itemsets generation. Some research works in maximal frequent itemsets mining are reported in [16] and [17]. However, there are only few algorithms adapted to nowadays distributed systems such as the grid. We can cite the GridDDM approach [18], which also mines maximal frequent itemsets from distributed datasets. It is based on a sequential maximal miner locally followed by a global mining phase.[1] Another approach for large scale association rules mining is proposed in [19]. It proposes a majority vote protocol acting in an asynchronous way over dynamic datasets. There exist some papers summarising the work that has been done in this area, such as [20] and [21] by Zaki. A relatively more recent overview is given in [22]. Although the authors consider a general survey, many aspects apply to the frequent itemsets mining.

Even some of these approaches were focusing on minimising the communication costs, most of them are under an implicit homogeneity hypothesis since they still need multiple synchronisation and communication phases between processes. Our aim is to propose and test a well-adapted approach dealing with constraints related to large distributed environments and their underlying tools. Indeed, we are designing and implementing the ADMIRE framework [23,24] which is a data-mining engine for the grid. It aims to provide grid-based data-mining techniques and implements higher knowledge map representations of mined data by both locally and globally. Similar grid-based projects and frameworks addressing this area include Knowledge Grid [25], Discovery Net [26], Grid Miner [27], and TeraGrid [28]. These frameworks aim to offer a high level abstractions and techniques for distributed data mining and knowledge extraction from data repositories and warehouses available on the grid. They often use basic grid mechanisms, mainly provided by existing grid environments, to build their specific knowledge discovery services. However, beyond their architecture and design, the data analysis and management policies, the data integration or placement approaches, or the underlying middleware and tools, the grid paradigm needs efficient and well-suited algorithmic approaches to optimize their performances, which is the main motivation of this work.

## 3 Problem definition

In this section, we define the frequent itemsets generation problem and its properties, and the underlying sequential method, i.e the Apriori algorithm. The proposed approach and the formalism that describes will be given in the next section.

3.1 Frequent itemsets mining

The frequent itemsets generation problem can be described as follows. Let $X = \{x_1, x_2, \ldots, x_n\}$ be a set of $n$ items, and let $W = \{t_1, t_2, \ldots, t_m\}$ be a set of $m$ transactions, where each transaction $t_i$ is a subset of $X$. An itemset $x$ of size $k$ is known as a $k$-itemset, and the support of $x$ is $\sum_{j=1}^{m}(1 : x \in t_j)$, i.e. the number of transactions in $W$ that have $x$ as a subset, denoted as $\sigma_x(W)$. The frequent itemset mining task is to find all $x \in W$ with support greater than a user-specified minimum value. A frequent itemset is maximal if it is not a subset of any other frequent itemset.

---

[1] Note that the GridDDM approach is not considered for comparison because it is not Apriori-based (which is the focus of this article).

As mentioned before, many existing frequent itemset generation algorithms, both sequential and distributed, are related to the Apriori algorithm [1]. Some sequential Apriori-based approaches in centralized datasets are reported in [2,4,5,29]. The Apriori algorithm exploits the observation that all subsets of a frequent itemset must be frequent. This algorithm does the following computation:

1. it starts with the generation of frequent 1-itemset, then iterates steps 2 and 3,
2. generate candidates at level $l$ from frequent itemsets at level $l-1$,
3. examine the data to determine whether the candidate sets meet the user-defined minimum support threshold. Remember the cases where the answer is positive, i.e. frequent $l$-itemsets, and go to the step 2.

3.2 Distributed frequent itemsets mining

The distribution aspect can be described as follows. Let $W$ be a dataset with $D$ transactions partitioned horizontally over $M$ nodes; $\{\varsigma_1, \varsigma_2, \ldots, \varsigma_M\}$. Let the size of the partition $\varsigma_i$ be $D_i$. Let $\sigma_x(W)$ and $\sigma_x(\varsigma_i)$ be the respective support of the itemset $x$ in $W$ and $\varsigma_i$. For a given minimum support threshold $s_W$, an itemset $x$ is *globally frequent* if $\sigma_x(W)$ is greater than $s_W \times D$, and is *locally frequent* at a node $N_i$ if its support $\sigma_x(\varsigma_i)$ is greater than $s_W \times D_i$. Here are two basic properties:

**Property 3.1** *A globally frequent itemset must be locally frequent in at least one node.*

*Proof* Let $x$ be an itemset. If $\sigma_x(\varsigma_i)$ is smaller than $s_W \times D_i$ for $i = 1, \ldots, M$, then $\sigma_x(W)$ is smaller than $s_W \times D$ (since $\sigma_x(W) = \sum_{i=1}^{M} \sigma_x(\varsigma_i)$ and $D = \sum_{i=1}^{M} D_i$), and $x$ cannot be globally frequent. Then, if $x$ is globally frequent, it must be locally frequent in at least one node $N_i$. □

**Property 3.2** *All subsets of a globally frequent itemset are globally frequent.*

*Proof* Let $x$ be an itemset, and let $x'$ be a subset of $x$. If $\sigma_{x'}(W)$ is smaller than $s_W \times D$, then $\sigma_x(W)$ is also smaller than $s_W \times D$ (since $\sigma_x(W) \leq \sigma_{x'}(W)$), and $x$ cannot be globally frequent. Then, if $x$ is globally frequent, all its subsets must be frequent. □

# 4 Algorithm

Basically, our approach computes the required size $k$ of frequent itemsets locally in each site without intermediate support counts exchange. All globally frequent itemsets are deduced at the end of the computation phase by considering:

1. at the first pass, the global support counts collection for frequent itemsets of size $k$ and all smaller frequent itemsets in each node that are not subsets of any greater size (maximal ones), and
2. subsets of locally frequent itemsets that fail the global frequency test in the following passes.

Note that a local pruning implicitly applies during these passes. Indeed, while generating subsets, some of them can become globally large by adding support counts of the different higher size itemsets from where they have been derived.

This distribution scheme is motivated by the performance behaviour of the basic generation task using Apriori, and the statement that is in widely distributed systems, such as the grid, it is highly recommended to avoid multiple synchronisation and communication steps as much as possible. Indeed, the overhead due to the communications can be very large for dense datasets or small minimum supports. The pseudo-code is presented in Algorithm 1. The loop **'for'** (in line 1) computes locally frequent itemsets in each of the $M$ nodes. $LL_i$ represents the locally frequent itemsets on a node $N_i$. The initial $LL_i$ sets are generated by the AprioriGen function (in line 2), then replaced (in the following iterations) by the subsets of the locally frequent itemsets that fail the global frequency (line 18). The $L_i$ set represents the itemsets that pass the global frequency test on a node $N_i$ (which is updated in line 17). Then, while the $LL_i$ set on $N_i$ is non empty (line 5), it is sent for remote support counts computation and collection (loops **'for'** in lines 6 and 9). The following section gives the analytical analysis of our approach.

**Algorithm 1.** GFM (Grid-based Frequent itemsets Mining).

**Input** : $\varsigma_i, i = 1, \ldots, M, s_W$ and $k$
**Output**: $L$, globally large itemsets of size 1 to $k$

```
1  for i = 1 to M do
2  |   AprioriGen(ς_i, k);
   end
4  for i = 1 to M do
5  |   while LL_i ≠ ∅ do
          //send locally frequent itemsets
6  |   |   for j = 1 to M (i ≠ j) do
7  |   |   |   Send(LL_i, j);
       |   end
       |   //remote computation
9  |   |   for j = 1 to M (i ≠ j) do
10 |   |   |   ReceiveRemoteSet(i, LL_j);
11 |   |   |   LocalSupport(LL_j);
12 |   |   |   SendSupportCounts(i, j);
       |   end
       |   //support counts reception
14 |   |   for j = 1 to M (i ≠ j) do
15 |   |   |   ReceiveRemoteSupports(j);
       |   end
       |   //compute globally large and deduce
       |   //subsets from globally failing itemsets
17 |   |   L_i += ComputeGloballyLarge();
18 |   |   LL_i = SubsetsGloballyFailled();
       |   end
   |   end
21 L = ⋃_{i=1}^{M} L_i;
```

### 4.1 Analytical analysis

This section presents the analytical study of this article. It has a performance analysis of the local Apriori task and the communication needs under a parameterised model, for both approaches: a typical distribution scheme, represented by the FDM algorithm, and the proposed approach. Recall that our aim is to show that intermediate global pruning steps are computationally inefficient in local nodes and affect the global system performance. This is related to the performance behaviour of the Apriori generation described here. This also must hold for other frequent itemsets mining techniques since it is also related to the datasets properties and/or the support thresholds.

The success rate at each level is the main factor determining the number of candidates at the next level. Besides, this latter parameter determines the amount of work that the algorithm will do. Interesting upper bound combinatorial analysis of this parameter is given in [30]. Experiments show that this bound is exact for some distribution, and that is close to the exact value in many cases. We will consider the upper bound of the amount of work done by the Apriori task at each level and each node. This is proportional to the global candidates at the level $l$ (denoted by $GC_{l,i}$, $i$ representing the computing node) for the FDM approach, and the local candidates at the level $l$ (denoted by $LC_{l,i}$ at a given node $N_i$) for the proposed approach. The main insight of our approach is that the set $GC_{l,i}$ obtained using a global pruning strategy is very close to the set $LC_{l,i}$ obtained only by local pruning. This leads to the fact that global pruning strategies (and the related multiple synchronisation steps) are performance constraining, and that this can be done in one phase at the end of the counting phases much more efficiently.

Consider now the notation of the set $GC_{l,i}$ as $GS_{l,i} + GFa_{l,i}$, which represents the sum of the number of itemsets that are success, and the number of the itemsets that fail the frequency test, at a given level $l$. This set also represents $LF_{l,i}$, the set of locally frequent itemsets at the level $l$. Consider the same notation for the set $LC_{l,i}$ as $LS_{l,i} + LFa_{l,i}$ associated with the proposed approach. Then, let $GC_{l+1,i} = GS_{l+1,i} + GFa_{l+1,i}$ be the number of candidates at the level $l + 1$. Now, consider the global view of a typical distributed approach, and $I_{l,i}$ the number of items involved at the level $l$ on the node $N_i$. On the one hand, by definition, each candidate on level $l + 1$ is associated with $l + 1$ frequent itemsets at level $l$, denoted by $GC_{l+1,i}$ $(l + 1)$. On the other hand, this number is bounded by $GS_{l,i}$ $(I_{l,i} - l)$. Indeed, there are at most $I_{l,i} - l$ candidates generation at the level $l + 1$ if the number of items that occurs at the level $l$ is $I_{l,i}$. Then we have:

$$GC_{l+1,i} \, (l+1) \leq GS_{l,i} \, (I_{l,i} - l)$$
$$GC_{l+1,i} \leq \frac{(I_{l,i} - l)}{l+1} \, GS_{l,i}$$

which gives

$$\sum_{l=0}^{I_{l,i}-1} \frac{(I_{l,i} - l)}{l+1} \, GS_{l,i}$$

as the upper bound of the amount of work. This can be associated with the global view of any distributed Apriori-based algorithm. However, we have usually $GS_{l,i} \leq LS_{l,i}$ (since $GC_{l,i} \leq LC_{l,i}$), then we introduce the factors $P_l$ and $P_{I_l}$ (related to items) and rewrite $GS_{l,i}$ as $P_l \, LS_{l,i}$. Thus, we can rewrite the sum above as:

$$\sum_{l=0}^{I_{l,i}-1} \frac{(P_{I_l} \, I_{l,i} - l)}{l+1} \, P_l \, LS_{l,i}$$

which represents local upper bounds related to both distributed approaches with $P_l$ and $P_{I_l}$ equal to 1 for the proposed approach. The lower bounds for $P_l$ and $P_{I_l}$ will be evaluated by experiments. The following models are mainly based on these amounts.

We also evoked the notion of critical or switch level which occurs in almost all experimental cases independently from the dataset types. Indeed, there is a value of $l$ such that the success rate is relatively high until this level, but lower and usually drops to zero afterwards.

To take into account, the variations that occur between various datasets, we can define a factor $F_l$ to quantify the changes in comparison to the bound defined before. If we use $l_c$ to be the critical level, $F_l$ is close to 1 if $l \leq l_c$, and close to 0 otherwise. The lower and higher bounds of $F_l$ in between, for the two cases, can be evaluated by experiments for a given dataset. However, this factor is a property of the dataset, and since the critical level is the same for both approaches for a given dataset, it will not be considered as a metric in the following models.

Now, in order to explain the network part of the proposed models, we present general schemes of both approaches. Basically, the FDM approach does the following computation:

*Step 1* Candidate sets generation, which are the locally frequent set at each node, using the Apriori algorithm,

*Step 2* Support counts exchange, which consists of two communication phases, and a computation phase for remote supports countings at each node, denoted by $RS_{l,i}$ at the level $l$ on the node $N_i$,

*Step 3* Generate the globally frequent itemsets at each node, of the current level, according to the collected remote support counts, followed by their broadcast. The steps 1 to 3 are repeated until reaching the requested size $k$. Note that an extra communication cost can occur at related to polling sites selection which is not considered for simplicity.

For the proposed approach, iterations are local rather than global since no global pruning is considered. The general scheme is as follows:

*Step 1* Apriori generation until reaching the requested size $k$,

*Step 2* A global support counts exchange phase. This consists of a few communication and computation passes for remote support counts collection, denoted by $RSg_{l,i}$. It generates globally frequent itemsets at each node using a top-down search.

As stated earlier, the network part is captured by the LogP model. This model is motivated by the current technological trends in distributed computing towards large grained networks. It has been shown to model accurately a variety of systems including grid-like environments. The main parameters of this model are; $L$ an upper bound on the latency, $o$ the overhead of handling a message, $g$ the gap (the reciprocal corresponds to the available bandwidth per processor), and P the number of modules or resources available. Using these metrics, if we use $k$ to be the requested level, the communication cost for the classical distribution is:

$$C_{\text{FDM}} = \sum_{l=1}^{k} \left( 2P \sum_{i=1}^{P-1} \left( GC_{l,i}\, g + L^2 + o \right) \right)$$

In the case of our approach, this cost is:

$$C_{\text{GFM}} = 2P \sum_{i=1}^{P-1} \left( LF_{k,i}\, g + L^2 + o \right) + \sum_{l=k-1}^{x} \left( 2P \sum_{i=1}^{P-1} \left( SF_{l,i}\, g + L^2 + o \right) \right)$$

where $SF_{l,i}$ represents subsets of global candidates that fail the frequency test at the earlier pass. Recall that $GC_{l,i}$ can be noted as $P_l\, LC_{l,i}$ using the factor $P_l$ introduced before, and that $LF_{k,i}$ represents also the set $GC_{k,i}$ generated locally. Then, the previous formalism leads to the following overall cost for the FDM distribution:

$$\sum_{i=1}^{P} \sum_{l=1}^{k} \left( \left( F_l \sum_{l=0}^{I_{l,i}-1} \frac{(P_{I_l} I_{l,i} - l)}{l+1} P_l \, LS_{l,i} \right) + RS_{l,i} \right) + C_{\text{FDM}} \qquad (1)$$

and for the proposed approach:

$$\sum_{i=1}^{P} \left( \sum_{l=1}^{k} \left( F_l \sum_{l=0}^{I_{l,i}-1} \frac{(I_{l,i} - l)}{l+1} LS_{l,i} \right) + \sum_{l=k}^{x} RSg_{l,i} \right) + C_{\text{GFM}} \qquad (2)$$

The upper and lower bounds of the factors $P_l$ and $P_{I_l}$ will be defined by experiments, as well as the upper bound of the amount of work at each level. We will evaluate the amount $x$ [used in $C_{\text{GFM}}$ and Eq. (2)], where $k - x$ represents the number of passes in GFM for the tested datasets (in comparison to $k$ passes for the FDM approach). We will also evaluate a lower bound for the difference factor between the amounts $GC_{l,i}$ and $SF_{l,i}$ related to support counts collections for the two approaches. This gives the amount of communication each algorithm will do. Also, since the factor $F_l$ is a property of the dataset and is appearing at the same level in the comparison, the models can be relaxed by ignoring it. Indeed, a given dataset will have the same properties and behaviour in both cases.

## 5 Experimentation setup

In this section, we briefly present the workflow concept, and the grid middleware (and its workflow manager) used in our implementations. Then, the proposed approach and the FDM algorithm are implemented, tested, and evaluated using this environment, and two widely used synthetic and real world datasets.

### 5.1 Workflow management

Several significant research works have been conducted in recent years to automate applications workflow management and execution on the grid. The concept of workflow or process arrangement is extremely important for distributed applications within the grid context. A large number of tools, with different and a large range of capabilities, have been implemented and used, including the Condor's DAGMan meta-scheduler [31], Askalon [32], CoG [33], YvetteML [34], GridAnt [35], among many others. Description languages for declaring the jobs composition have also been proposed, and many of them have an XML-based modeling. The architecture is usually composed of a user interface or language tools and a workflow execution engine, which controls the execution on the grid. In the following, we briefly present the Condor system and its workflow manager used in our implementation.

### 5.2 Condor/DAGMan

The Condor system is a distributed batch system providing a job management mechanism, resource monitoring and management, scheduling and priority schemes. The Condor system provides a ClassAds mechanism for matching resource requests and offers a checkpointing and migration mechanisms. It provides also job management capabilities for the grid through

Condor-G (using the Globus Toolkit) and Condor-C, which allows jobs to be moved between machines job queues.

DAGMan is a directed acyclic graph representation manager, which allows the user to express dependencies between Condor jobs. It allows the user to list the jobs to be done with constraints on the order through several description files for the DAG and the jobs within the task graph. It also provides fault–tolerant capabilities allowing to resume a workflow where it was left off, in the case of a crash or other failures. However, the scripting language required by DAGMan is not flexible since every job in the DAG has to have its own submit description file.

5.3 Experimentation platform

The experiment platform is a cluster of workstations connected by a Gigabit Ethernet. It consists of 312 bi-processors AMD-64 opteron 2 GHz and 2 GB of memory. This cluster is one of the nine nodes of the Grid'5000 platform [36] which is a large scale computing environment for grid research. It provides a set of control and monitoring tools allowing users to make reservations, and configure a specific environment, i.e. required software packages and/or operating system.

## 6 Performance evaluations

In this section, we will present bounds estimation and experimental evaluations of the described approaches. We used a sample from the PUMS census dataset and a generated synthetic dataset using the IBM Quest code. This latter models supermarket basket data. It has been used in several frequent itemsets generation studies [3,15,37], etc.

It is important to note that performance estimations of any distributed system depends on the current experiments conditions, and a range of overheads related to the middleware and the execution engine. This can affect the comparison. As a consequence, we will not include some overheads in the reported execution times to keep the comparison uniform, such as the job preparation and submission overhead, which are variable, and sometimes very excessive under some grid middleware. Indeed, while the performance problem is usually a complex scheduling issue in the case of complex applications, previous large scale distributed implementations, including a distributed variance-based clustering algorithm [38], showed that it is more related to the underlying tool in the grid since basic jobs were mostly organized in large parallel loops. Furthermore, a simple workflow executed locally (in a laptop, Genuine Intel Centrino Duo 2 GHz and 2 GB memory) and containing only two small jobs under the Condor/DAGMan system, shows that the preparation process takes in average about 295 s (about 5 min). This represents the time interval between the workflow launching and the first job submission. However, this does not seem to be related to the size of the DAG and can be variable.

6.1 Bounds estimation

In this section, we compute the bounds of the factors described in the analytical study, and representing the basis of the comparison. Based on Eqs. (1) and (2) in Sect. 3.3, we will show that the estimated computational gain in the local Apriori generation process for the FDM version, resulting from lower candidate sets sizes, is negligible in comparison to the generation cost of the proposed approach. Also, the top-down global frequency generation is much less costly in terms of communications and synchronisations, and is more computationally efficient.

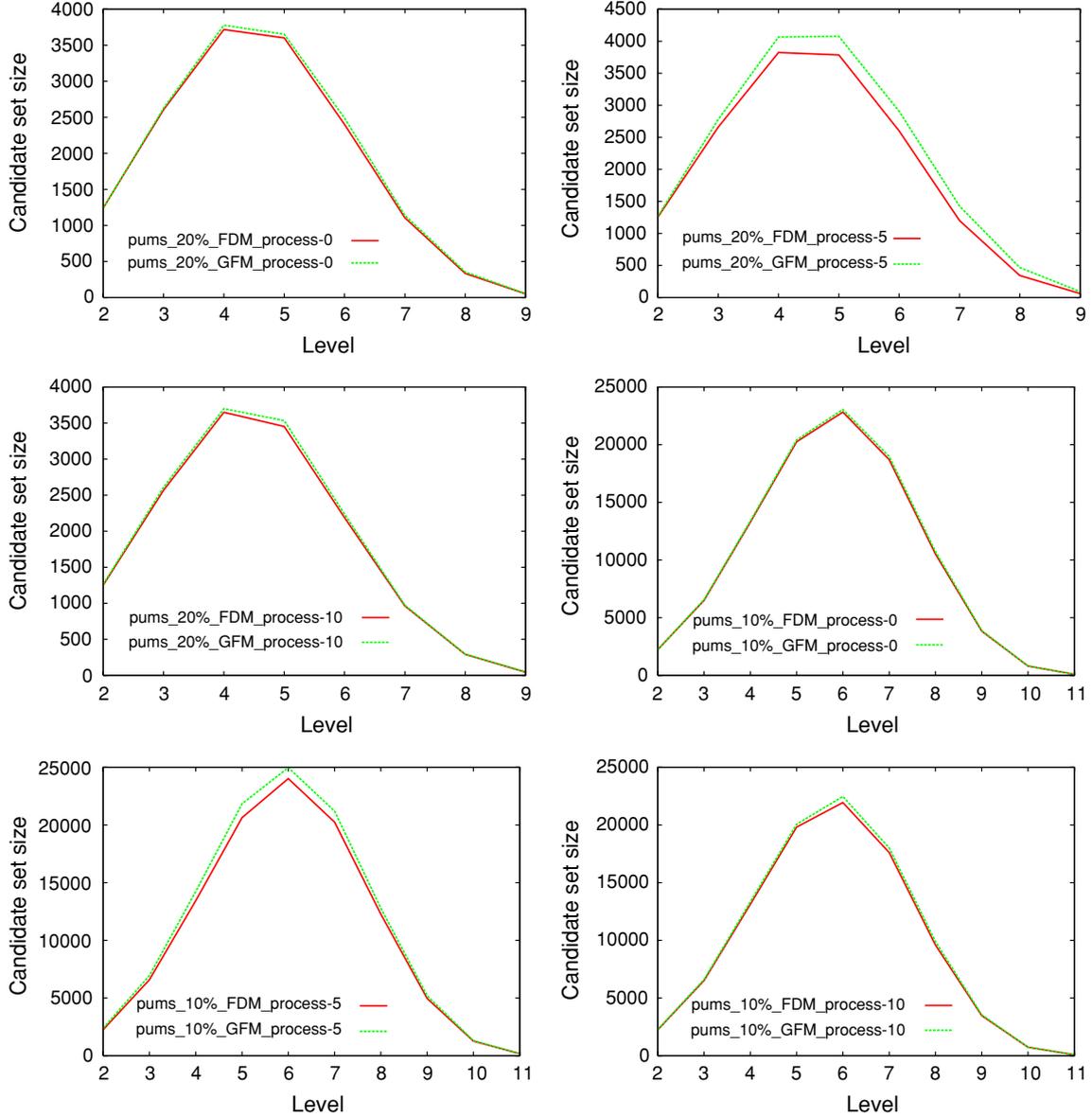

**Fig. 1** Generated candidate sets using both approaches on different processes

In the following, we will give the experimental bounds of the factors $P_l$ and $P_{I_l}$ for the considered datasets. Besides, we will estimate the communication costs at each level on the basis of the observed sizes of the generated candidate sets, i.e. the sets $GC_{l,i}$ and $SF_{l,i}$ for FDM and GFM, respectively. We will give experimental estimations of the gain in the computing time in the local Apriori generation process (related to the factor $P_l$). We will also discuss the process related to remote support counts computations.

Figure 1 shows plots of different candidate sets on different nodes using various support thresholds, for both datasets. The lower bound of the factor $P_l$ is 0.78. This value is close to 1 in most cases, and its average value is 0.93. We will consider the same factor for both the candidate sets sizes and success rates since it is almost the same. The factor $P_{I_l}$ follows the same behaviour with a lower bound equal to 0.87, and the average value is 0.94. The resulting difference in the experimental computing cost of the local Apriori generation is insignificant (of the order of seconds in all cases). Comparisons of the average communication needs ($C_{FDM}$ and $C_{GFM}$ defined in Sect. 3.3), related to the global strategies of both approaches are given in Fig. 2. It shows that the global strategy of the GFM approach is much less consuming in terms of communication compared to the FDM approach. Indeed, GFM needs only two communication passes for both datasets. The gain factor reaches 82%, and the lowest rate is 8% while considering the worst execution cases of the GFM algorithm related to small values of $k$ for the synthetic dataset. Note that the input/output parameter was not considered in the analytical model for simplicity. This is more costly in the case of the FDM approach since the proposed approach generates less important overall sets for remote support counts collection. The next section will give details about the conducted tests and the overall execution times.

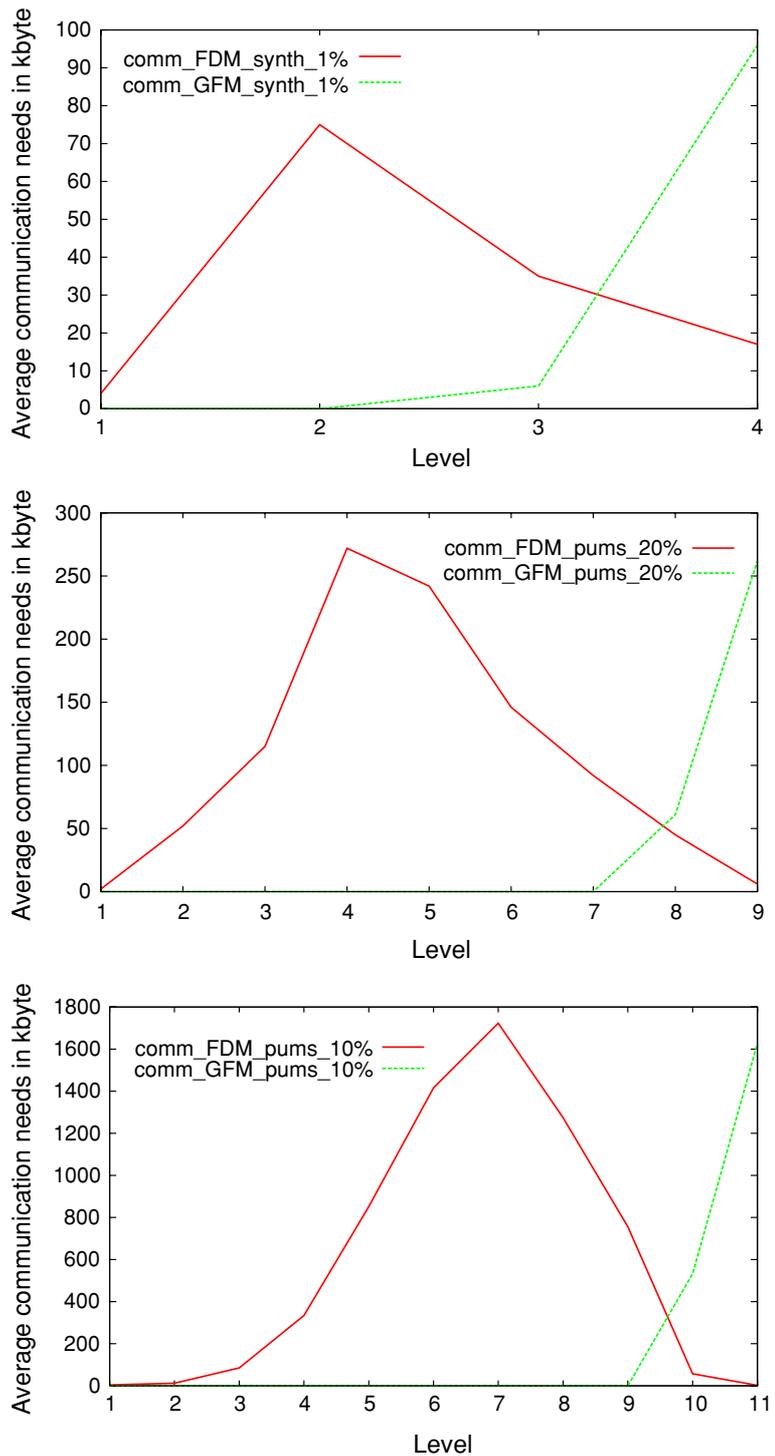

**Fig. 2** Average communication needs per process for both approaches

Bounds evaluation presented in this section specifically show that: (1) distributed implementations of the Apriori algorithm do not need global pruning strategies, and therefore,

**Table 1** Average execution times, FDM versus GFM—synthetic dataset

| Support (%) | Ratio | Size (×10⁶) | FDM | GFM | Factor (%) |
|---|---|---|---|---|---|
| 1 | 1:1 | 0.5 | 188 | 172 | 8 |
| 2 | 1:1 | 0.5 | 36 | 31 | 13 |

(2) classical distribution schemes are less more efficient than the adopted global strategy in our approach, starting from the requested size and using a top-down search, which in our case led to only two communication passes per execution. Note that remote support counts computations is a very costly process in FDM, specially in lower levels where the success rates are high. This process was reduced to the minimum in our approach since only two remote computation passes are required.

### 6.2 Experimental results

We present now the execution setup of the two implementations, GFM and FDM. The dataset collections are composed of up to $10^6$ transactions distributed by size ratios of 1:1 to 1:10.[2] This local execution can be block-based, i.e. the local portions can be partitioned into blocks beforehand in the presence of out-of-core datasets to avoid any thrashing effect, which can drastically degrade the performance. The support thresholds are set from 10 to 20% for the PUMS dataset, and 1 and 2% for the synthetic dataset. The average transaction size is 20 for the synthetic dataset, and 30 for the census dataset. The size of blocks can be different depending on the available memory, the threshold support, the number of items, etc. However, since the testbed was dedicated, and depending on the nature of the used datasets, these amounts were fixed (as upper bound sizes) to 10,000 transactions for the synthetic dataset and about 5,000 for the census dataset.

First, we perform tests with homogeneous distributions, i.e. size ratio equal to 1:1. Note that our approach consists usually of few parallel loops, with at most two communication passes in all tested cases. Whereas, if $k$ is the requested size of frequent itemsets, the FDM approach consists of $k$ communication steps for support requests, $k + 1$ parallel activities for Apriori generation processes, and $k$ parallel activities for remote supports computation. For this first set of tests, the datasets size is $0.5 \times 10^6$ transactions, distributed over 50–100 CPUs. In the DAGMan workflow framework, each job is represented by a submission file and all dependencies are represented by a DAG file.

We consider the generation of frequent itemsets to the maximal size for both datasets. The average processing times for the described sizes are given in Tables 1 and 2. These show a better performance by up to 79% of the computing time for the GFM version. Note that, in addition of the multiple communication and synchronisation steps in the FDM version, the remote support computations at the different exchange stages are also quite computationally expensive, and reach 21% of the whole processing time presented here. This highly constrains the performance, specially in lower generation levels with high success rates. The gain factor is less important in the case of the synthetic dataset (up to 13%) because of less generation stages, i.e small maximal frequent itemsets size. Indeed, $k$ reached only 4 for the synthetic dataset, against 9–11 (depending in the support threshold) for the PUMS dataset. In addition, depending on the underlying tools used for the implementation, and taking into account all kinds of overheads in the grid hierarchy, this will increase the gain of the proposed approach since less communication and synchronisation stages are required, in addition to lower number of jobs in the DAG. We will briefly discuss the overhead issue later on.

---
[2] This means that local portions can be up to ten times larger in some nodes.

**Table 2** Average execution times FDM versus GFM—PUMS dataset

| Support (%) | Ratio | Size ($\times 10^6$) | FDM | GFM | Factor (%) |
|---|---|---|---|---|---|
| 10 | 1:1 | 0.5 | 2,305 | 827 | 64 |
| 20 | 1:1 | 0.5 | 857 | 180 | 79 |

**Table 3** Average execution times with a heterogeneous distribution—synthetic datasets

| Support (%) | Ratio | Size ($\times 10^6$) | FDM | GFM | Factor (%) |
|---|---|---|---|---|---|
| 1 | 1:5 | 0.5 | 586 | 523 | 10 |
| 2 | 1:5 | 0.5 | 163 | 149 | 8 |
| 1 | 1:10 | 1 | 1,342 | 1,181 | 12 |
| 2 | 1:10 | 1 | 314 | 286 | 9 |

**Table 4** Average execution times with a heterogeneous distribution—PUMS dataset

| Support (%) | Ratio | Size ($\times 10^6$) | FDM | GFM | Factor (%) |
|---|---|---|---|---|---|
| 10 | 1:5 | 0.5 | 3,917 | 1,371 | 65 |
| 20 | 1:5 | 0.5 | 2,297 | 508 | 77 |
| 10 | 1:10 | 1 | 9,658 | 3,090 | 68 |
| 20 | 1:10 | 1 | 5,340 | 1,026 | 81 |

In the following tests, we used a size ratio up to 1:10 for the datasets distribution. Indeed, FDM behaves badly in this case since the cost related to the waiting time, at each synchronisation step, for the slowest or the most overloaded node can be of factor $k - 1$ between the two versions. The local dataset portion sizes are ranging between 10,000 and 100,000 transactions (sizes generated randomly), with similar sizes of blocks as described earlier. The average processing times to reach the maximal size of frequent itemsets are given in Tables 3 and 4. For these tests, the GFM algorithm improves the execution performance by 8–81% compared to the FDM implementation. On the one hand, the results show that the gain factor is more important in the case of more important values of the generation level $k$ as quoted before. On the other hand, uneven dataset distributions do not greatly increase the gain factor in our case because of the test bed homogeneity.

6.3 Discussion

Our implemented approach within a workflow environment for the grid aims to reduce to the minimum task synchronisations and data communications. The reasoning behind this approach is based on the inherent nature of the generation task itself described in this article. The comparison between the presented approaches shows the effectiveness of the GFM algorithm and give good targets for future versions and evaluations.

Another important aspect in the design of grid-based algorithms is the scalability issue, i.e., how to use the increasing amounts of processing resources in a memory and bandwidth constrained environment? Most existing approaches for frequent itemsets generation need to be adapted locally to tackle the memory constraint issues, and do not take into account the underlying environment nature. The tests show that in the case of increasing sizes and number of processes, the data structure related to the large amount of remote supports exchange is memory constraining and need explicit input/output management for local intermediate countings. This is more pronounced in lower levels. The proposed approach is more scalable in these cases since the counting iterations are local and do not take into account the large number of remote condidate sets. Current large distributed systems and the related constraints put then pressure on the design of effective distributed algorithms. Our approach aims to tackle some of these issues based on the main features of the underlying task using a performance model and different types of datasets.

As for the execution overheads, important latencies related to the underlying tool were noticed, such as the job preparation and submission or task scheduling. These are of prime importance and can be the first source of efficiency loss, in addition of affecting comparison results. However, we did remove them from the reported computing times, which do not include these specific middleware related overheads. Note that latencies related to job preparation and submission were on average up to 13 times more important in the FDM approach compared to the GFM implementation. This was expected since there are much more computational stages, jobs, and synchronisations in the FDM implementation. Our evaluation took then into account the time related to each parallel loop for the two versions including its communication needs. The related overheads can reach up to 23% of the overall execution time, depending on the number of levels and the number and size of jobs. The severity of some cases, however, highlights the need of more consideration of this issue from the grid community. Details about the hierarchy of these overheads will be more considered in future work.

## 7 Conclusion

In this article, we presented a performance study of distributed Apriori-like frequent itemsets mining, and proposed a well-adapted distributed approach which leads to a substantial improvement in the efficiency of its implementation. On the one hand, we evoked the need of efficient distributed approaches well-suited for loosely coupled environments such as the grid. On the other hand, based on the Apriori algorithm properties, we showed that the intermediate communication steps are computationally inefficient since the global prunings do not bring enough useful information, which consequently greatly affects the global performance. Support counts collections passes are then much more efficient at the end of local counting phases. This is also highly suitable in the case of uneven dataset distribution and/or platform heterogeneity.

The proposed approach is basically intended to limit synchronisation and communication overheads, in addition to the underlying grid tools overheads. Comparisons to a typical Apriori-like distributed approach show its effectiveness. This also attests that distributed implementations on the grid have an essential need to avoid multiple communication and synchronisation steps as much as possible.